# Learning to Optimize via Wasserstein Deep Inverse Optimal Control

Yichen Wang, Le Song, and Hongyuan Zha

College of Computing, Georgia Institute of Technology
yichen.wang@gatech.edu, {lsong, zha}@cc.gatech.edu

**Abstract**

We study the inverse optimal control problem in social sciences: we aim at learning a user's true cost function from the observed temporal behavior. In contrast to traditional phenomenological works that aim to learn a generative model to fit the behavioral data, we propose a novel variational principle and treat user as a reinforcement learning algorithm, which acts by optimizing his cost function. We first propose a unified KL framework that generalizes existing maximum entropy inverse optimal control methods. We further propose a two-step Wasserstein inverse optimal control framework. In the first step, we compute the optimal measure with a novel mass transport equation. In the second step, we formulate the learning problem as a generative adversarial network. In two real world experiments — recommender systems and social networks, we show that our framework obtains significant performance gains over both existing inverse optimal control methods and point process based generative models.

## 1 Introduction

Internet is transforming our lives: now people are relying increasingly on Internet to communicate with their friends, participate in discussions, and stay in touch with the world. Millions of people around the world generate large-scale temporal behavior data from various domains, such as social platforms, healthcare systems, and online service websites. As life becomes increasingly digital — behavior and decisions are represented as bits, sets, events, and time series — the key challenge to tackle is: How to make these digital platforms more useful and engaging for online users and the entire society? To tackle this challenge, mathematical modeling frameworks are necessary to properly analyze the complex behavior of users, and to apply the resulting insights in the real world.

Existing works in computational social sciences and user modeling typically start from the *phenomenological* principle: developing and fitting generative models to describe the behavioral data. For example, point process based models have been applied in diverse applications, such as information diffusion [5, 11, 26, 27], recommender systems [6], and evolutionary networks [26]. However, these generative models might not be expressive enough, since users typically are optimizing utility functions to make decisions and take actions. For example, in social networks, different users may have different cost functions to consider when they make posts. Some users might be easily influenced by neighbors and peers, hence the majority of their posts could be the retweets of friends' post, and their tweets could also share similar topics of friends' posts. On the other hand, other users might be more self-centric, and mostly posts about their own life events or personal thoughts, regardless of their friends' posts. These complex reasoning and optimizing mechanism from users are not captured in these generative models.

Instead of the phenomenological perspective, we propose a new *variational* (optimization) principle: we treat the behavior data as a result of the optimization scheme executed by the user. The dynamic system for a user includes its intrinsic baseline dynamics, cost function, and control policy. The control policy is obtained as a solution to an optimization scheme. Our goal is *learning to optimize*, *i.e.*, learning the cost function of users' optimization scheme. Discovering users' cost function is important: it helps service platforms understand the causality in users' behavior, and provide personalized services to improve user experience. In this paper, we are interested in the inverse optimal control/reinforcement learning problem. Formally, given samples of behavior trajectories, we treat users as an RL algorithm that solves an optimization problem to make decisions/policies. Our goal is to infer the cost function that drives the observed behavior.

The simplest approach to inverse optimal control (IOC) is behavior cloning [19], which learns a policy as a supervised learning problem from expert/optimal trajectories. However, this method is only efficient with large amount of data and suffers from the issue of cascading errors caused by covariate shift [18, 20, 22, 23, 24]. Based on the seminar work of maximum entropy IOC [4, 14, 33], IOC has been successful in many domains, including predicting taxi drivers [33] and robots [21]. However, IOC has been expensive to run, requiring reinfocement learning in an inner loop, recently [3, 9, 16, 30] proposed methods that scales to large environment using neural networks. Instead of learning cost functions, [12, 13] directly learn the policy using policy optimization or generative adversarial network. However, most of these works are based on the maximum entropy IOC method, which is limited by the KL divergence term in the optimization scheme.

Social science setting also introduces new challenges to IOC. Existing works are mostly based on Markov Decision Processes (MDPs). However, Online user activity modeling requires us to consider more advanced processes, such as Hawkes processes for long term memory and mutual exciting phenomena in social interactions, survival processes [1] for self-terminating behavior in influence propagation and link creation, and node birth processes for evolving networks [25]. For these stochastic processed, the probability distribution of trajectories is not as easy to calculate as in MDPs [12]. Thus, these existing works are not directly applicable to our problem, and new theories and algorithms are in urgent need to be developed. In summary, we make the following contributions.

- We propose a KL inverse optimal control framework, which is applicable to any stochastic processes and not limited to MDPs. Our framework consists of two steps. In the first step, we derive a general formulation of the parameterized optimal measure, which is in the form of a Radon-Nikodym derivative. In the second step, we infer the cost function using variational inference. We further show that this framework provides a principled and unified view of existing works in IOC.

- We design a novel Wasserstein inverse optimal control framework by using Wasserstein distance in these two steps instead of the KL divergence. In the first step, we explicitly characterize the optimal measure by solving a mass transport problem. In the second step, we infer the cost function using a novel generative adversarial networks.

- We present a novel variational principle to understand users' behavior — we treat users as reinforcement learning agents that aim to optimize their cost functions, which is in sharp contrast to existing works that are purely focused on fitting generative models to the observed data. In real world experiments, we show that our methods outperform the state of the art IOC algorithms as well as generative models in learning users' temporal behavior.

# 2 KL deep inverse optimal control

In this section, we present our KL deep inverse optimal control framework. We provide a novel view of IOC problem from a two-step procedure. In the first step, we find the optimal measure with a KL regularization; in the second step, we use variational inference to compute the cost function.

## 2.1 Finding the optimal measure

Users in social platforms make decisions by optimizing their cost functions. Let $c(\theta, \tau) = \sum_t c_\theta(x_t)$ be the user's state cost function, where $\theta$ is the unknown parameter of interest. Let $\tau = \{x_1, \cdots, x_K\}$ be the trajectory of the observed stochastic process $x_t$. Note our method is applicable to general stochastic processes, that includes the commonly used MDPs, point processes, and stochastic differential equations. Let $\mathbb{P}$ be the *baseline/uncontrolled* probability measure (distribution) induced by the baseline behavior of users without any optimization, and $\mathbb{Q}$ be the *controlled* measure induced by the user that uses some policy $\pi$ via optimization.

With the KL distance as the regularization, we propose the following optimization problem to compute an optimal measure that minimizes the expected cost function:

$$\boxed{\min_{\mathbb{Q}} \Big[ \underset{\text{state cost}}{\underset{\uparrow}{\mathbb{E}_{\tau \sim \mathbb{Q}}[c(\theta, \tau)]}} + \underset{\text{control cost}}{\underset{\uparrow}{\gamma D_{KL}(\mathbb{Q}||\mathbb{P})}} \Big]} \tag{1}$$

The rationale behind this formulation is two fold. First, the user tries to find the optimal policy (equivalent, the optimal measure) that minimizes his state cost. Second, he minimizes his control cost to change behavior. For example, the control cost can be in the form of money, and time to change the user's original behavior. We use the KL divergence to measure the divergence between the baseline measure $\mathbb{P}$ and controlled measure $\mathbb{Q}$, which evaluates the effort to change from $\mathbb{P}$ to $\mathbb{Q}$. The trade-off parameter $\gamma$ controls the weight between the state and control cost.

Using the property of the KL divergence, we can obtain an closed form solution to (1), and the optimal measure $\mathbb{Q}^*_\theta$ is expressed as follows (Appendix B contains derivations):

$$\boxed{\frac{\mathrm{d}\mathbb{Q}^*_\theta}{\mathrm{d}\mathbb{P}}(\tau) = \frac{\exp(-c(\theta, \tau)/\gamma)}{\mathbb{E}_{\tau \sim \mathbb{P}}[\exp(-c(\theta, \tau)/\gamma)]}} \tag{2}$$

The term $\frac{\mathrm{d}\mathbb{Q}^*_\theta}{\mathrm{d}\mathbb{P}}$ is called the Radon-Nikodym derivative [7], it can be viewed as the relative density function between $\mathbb{Q}^*_\theta$ and $\mathbb{P}$, it is the probability density of observing a trajectory $\tau$. This expression is intuitive: if a trajectory $\tau$ has low state cost, then $\frac{\mathrm{d}\mathbb{Q}^*_\theta}{\mathrm{d}\mathbb{P}}$ is large. This means that this trajectory is likely to be sampled from $\mathbb{Q}^*_\theta$. Next, we will use this expression to learn the cost function of users.

## 2.2 Finding the state cost

Given the observed optimal trajectories $\{\tau_i\}$ of the user, in this second step, we maximize the log-likelihood of these trajectories to learn the parameter of the cost function: $\max_\theta \sum_j \log \frac{\mathrm{d}\mathbb{Q}^*_\theta}{\mathrm{d}\mathbb{P}}(\tau_i)$.

Using the definition of KL divergence, one can show that this MLE estimation problem is equivalent to minimizing the following divergence between the observed empirical measure $\widetilde{\mathbb{Q}}$ and $\mathbb{Q}^*_\theta$:

$$\boxed{\min_\theta D_{KL}(\widetilde{\mathbb{Q}}||\mathbb{Q}^*_\theta)} \tag{3}$$

where the empirical probability measure $\widetilde{\mathbb{Q}} := \frac{1}{n}\sum_i \delta_{\tau_i}$ is the probability measure induced by the observed optimal trajectories $\{\tau_i\}$. Note that the KL distance is not symmetric and the order of two measures could not be reversed — this formulation requires that the empirical measure $\widetilde{\mathbb{Q}}$ have full support of $\mathbb{Q}^*_\theta$, which is a too strong assumption and might not be realistic in real world scenarios.

## 2.3 Problem with the KL divergence

In summary, our KL IOC framework uses the KL divergence as regularization in the first step and as the objective function in the second step. One can solve (3) using GANs. However, the problem with the KL divergence metric is that if $\widetilde{\mathbb{Q}}$ and $\mathbb{Q}^*_\theta$ are disjoint or lie in low dimensional manifolds, then this leads to an unreliable training of the generator if we use the KL divergence or its variant JS divergence [2]. This is likely to happen in social science data, where the users behavior are complex and $\widetilde{\mathbb{Q}}$ can have arbitrary support.

To overcome this limitation, next we propose to use the Wasserstein distance as the metric both in finding the optimal measure and the cost. There are also significantly new challenges in using the Wasserstein distance metric, which will be discussed and addressed in the next section.

# 3 Wasserstein deep inverse optimal control

In this section, we propose our framework based on Wasserstein distance for IOC. We first provide some background on the Wasserstein distance and then present our two-step framework.

## 3.1 Wasserstein distance

The Wasserstein distance is defined as

$$D_W(\mathbb{Q}||\mathbb{P}) = \inf_{\Pi \in u(\mathbb{Q},\mathbb{P})} \mathbb{E}_{(x,y)\sim\Pi} d(x,y) \tag{4}$$

where $u(\mathbb{P},\mathbb{Q})$ denotes the set of all joint probability distributions $\Pi(x,y)$ whose marginals are respectively $\mathbb{P}$ and $\mathbb{Q}$. Intuitively, $\Pi(x,y)$ indicates how much mass must be transported from $x$ to $y$ in order to transform the distributions $\mathbb{P}$ and $\mathbb{Q}$. This distance is the cost of the optimal transport plan. $d(x,y) = \|x-y\|$. The dual form of the Wasserstein distance is defined as

$$D_W(\mathbb{Q}||\mathbb{P}) = \sup_{\|f\|_L \leqslant 1} \Big( \mathbb{E}_{\tau\sim\mathbb{Q}}[f(\tau)] - \mathbb{E}_{\tau\sim\mathbb{P}}[f(\tau)] \Big)$$

where $\|f\|_L \leqslant 1$ means that the function $f(\cdot)$ is Lipschitz continuous with Lipschitz constant of 1.

## 3.2 Finding the optimal measure in Wasserstein distance

We use the Wasserstein distance as the regularization for the control cost, and propose the following objective function.

$$\boxed{\min_{\mathbb{Q}} \Big[ \mathbb{E}_{\tau\sim\mathbb{Q}}[c(\theta,\tau)] + \gamma D_W(\mathbb{Q}||\mathbb{P}) \Big]} \tag{5}$$

Different from the KL case, there is no closed form solution that can express the optimal measure $\mathbb{Q}^*_\theta$ as a function of $\mathbb{P}$. However, if we choose $\mathbb{P}$ as the empirical distribution. Then optimal $\mathbb{Q}_\theta$ can be computed in closed form as an empirical distribution, as stated in the following theorem.

**Theorem 1** (Optimal measure as a mass transport)**.** *Let $\mathbb{P}$ be the empirical distribution defined as $\mathbb{P} = \frac{1}{n}\sum_i \delta_{\tau_i^0}$, where $\tau_i^0$ is the uncontrolled baseline trajectory of a stochastic system. Then the optimal $\mathbb{Q}_\theta$ for the optimization problem has the following structure:*

$$\boxed{\mathbb{Q}^*_\theta = \frac{1}{n}\sum_{i=1}^{n} \delta_{\tau_i^*(\theta)}} \tag{6}$$

*where $\tau_i^*$ is computed by solving the following optimization problem*

$$\tau_i^*(\theta) = \operatorname*{argmin}_{\tau\in\mathbb{R}^K} \{c(\theta,\tau) + \gamma d(\tau_i^0,\tau)\} \tag{7}$$

**Proof sketch**. We will use the definition of Wasserstein distance in (4) to simplify the objective function in (5). Since $\mathbb{P}$ is an empirical measure, the optimal $\mathbb{Q}^*_\theta$ should also be an empirical measure. Appendix A contains details of the proof.

**Mass transport interpretation**. Theorem 1 presents a mass transport between two empirical measures: it shows the mechanism to transport each trajectory in the baseline empirical measure $\mathbb{P}$ to the optimal trajectory in the optimal measure $\mathbb{Q}^*_\theta$. This mass transport mechanism is also intuitive: the new trajectory $\tau_i^*(\theta)$ is the optimal one that minimizes the combination of state cost and control cost, which is the distance to the baseline trajectory $\tau_i^0$. Compared with the expression in (2), though an analytic expression is not present, we are still able to characterize the optimal measure.

### 3.3 Finding the state cost in Wasserstein distance

Given the optimal measure $\mathbb{Q}_\theta^*$, in the second step, we minimize the Wasserstein distance between the empirical measure $\widetilde{\mathbb{Q}}$ and $\mathbb{Q}_\theta^*$ as follows:

$$\boxed{\min_\theta D_W(\tilde{\mathbb{Q}} \| \mathbb{Q}_\theta^*)} \tag{8}$$

This problem can be re-formulated as the following optimization problem using the definition of the Wasserstein distance:

$$\min_\theta \sup_{\|f_w\|_L \leqslant 1} \Big( \mathbb{E}_{\tilde{\mathbb{Q}}}[f_w(\tau)] - \mathbb{E}_{\mathbb{Q}_\theta^*}[f_w(\tau)] \Big)$$

Our formulation is similar to the Wasserstein GAN [2]. However, the key difference is that we directly optimize $\mathbb{Q}_\theta^*$, which is an empirical measure, while [2] assumes that the measure $\mathbb{Q}_\theta^*$ is absolute continuous with respect to a parameterized reference measure. Hence the algorithm in [2] is not applicable to our problem.

To solve our problem, we use neural networks to parameterize the discriminator $f_w$ and the cost function $c(\theta, \tau)$. Note that the cost function uniquely determines the generator $\mathbb{Q}_\theta^*$. To learn the parameter $w, \theta$, we need to compute the gradients. Computing the gradient with respect to $w$ is standard, and the challenges lie in computing the gradient with respect to $\theta$. Next, we will discuss this gradient in detail.

**Gradient with respect to $\theta$.** We first define a new function $L$ as follows

$$L(w, \theta) := \mathbb{E}_{\mathbb{Q}_\theta}[f_w(\tau)] = \frac{1}{n} \sum_i f(w, \tau_i^*(\theta))$$

where $\tau_i^*(\theta)$ is defined in (7) and the second equality comes from the definition of $\mathbb{Q}_\theta^*$ in (6). Taking the gradient wrt $\theta$ yields:

$$\frac{\partial L(w, \theta)}{\partial \theta} = \frac{1}{n} \sum_i \frac{\partial f(w, \tau_i^*)}{\partial \tau_i^*} \frac{\partial \tau_i^*}{\partial \theta}$$

Computing the partial gradient $\frac{\partial \tau_i^*}{\partial \theta}$ is challenging, since $\tau_i^*$ itself is the optimizer of (7) that involves $\theta$ in the optimization. Hence computing $\theta$ is essentially a bi-level optimization problem.

We present the following theorem for computing this gradient.

**Theorem 2.** *The gradient $\frac{\partial \tau_i^*}{\partial \theta}$ has the following form:*

$$\frac{\partial \tau_i^*}{\partial \theta} = -\left[ \frac{\partial g(\theta, \tau_i^*)}{\partial \tau \tau} \right]^{-1} \frac{\partial g(\theta, \tau_i^*)}{\partial \theta \tau}$$

*where $g(\theta, \tau) := c(\theta, \tau) + \gamma d(\tau_i^0, \tau)$.*

This equation can be proved using chain rule and the fact that $\tau_i^*$ is the minimizer of (7). Appendix C contains details on the proof. With this gradient expression, we can efficiently compute the parameter $\theta$ for the cost function.

## 4 Comparison with existing works

In this section, we compare our two frameworks with three representative methods in inverse optimal control.

### 4.1 Maximum entropy IRL

We show that maximum entropy inverse reinforcement learning (MaxEnt) [32, 33] is a special case of our KL framework in Section 2. This method is the foundation of the state of the art [9, 12, 13]. MaxEnt propose to find a distribution that maximizes the probability of observed trajectories within the family

$$\frac{1}{Z} \exp(-c(\theta, \tau))$$

where $Z$ is the normalization constant. Compared with our method, we can see that this trajectory probability distribution are identical with our Radon-Nikodym derivative in (2) when *the uncontrolled dynamics* $\mathbb{P}$ *is uniform*. Hence MaxEnt is an inverse method with uniform uncontrolled dynamics. This restriction to uniform dynamics could be problematic in modeling users' temporal behavior in social sciences. The uncontrolled baseline user dynamics is typically generated by point processes and SDEs [5, 6, 11, 26, 27, 28, 29] and the probability distribution of these stochastic processes are much more complicated than a uniform distribution. Moreover, our KL framework directly shows the mathematical reasoning for choosing this specific form of probability distribution, which is nonexistent in previous works.

### 4.2 Guided cost learning

In the guided cost learning IOC work by Finn et al [9], the uncontrolled measure $\mathbb{P}$ is still assumed to be uniform. Finn et al proposed to learn another background distribution $\mathbb{Q}$ which generates the trajectories. Note that the objective function $\min_\theta D_{KL}(\widetilde{\mathbb{Q}}||\mathbb{Q}_\theta^*)$ in (3) can be expressed as

$$\min_\theta \sum_i c(\theta, \tau_i) + \log(\mathbb{E}_\mathbb{Q}[\exp(-c(\theta,\tau))\frac{\mathrm{d}\mathbb{P}}{\mathrm{d}\mathbb{Q}}]) \tag{9}$$

To learn $\mathbb{Q}$, this work also minimizes the KL divergence with respect to $\mathbb{Q}_\theta^*$:

$$\min_{\mathbb{Q}} D_{KL}(\mathbb{Q}||\mathbb{Q}_\theta^*) \tag{10}$$

The idea of guided cost learning suggests that one can iteratively update $\mathbb{Q}$ such that it is close to $\mathbb{Q}_\theta$ by minimizing the KL divergence $D_{KL}(\mathbb{Q}||\mathbb{Q}_\theta^*)$. In the follow-up work [8], they showed that this formulation is equivalent to a GAN [10] where $\mathbb{Q}$ is the generator and the optimal discriminator is defined as $D_\theta(x) = \frac{\mathrm{d}\mathbb{Q}_\theta/\mathrm{d}\mathbb{Q}}{\mathrm{d}\mathbb{Q}_\theta/\mathrm{d}\mathbb{Q}+1}$. Therefore, optimizing $\theta$ in (9) is equivalent to optimizing the discriminator $D_\theta$ and optimizing (10) is equivalent to optimizing the generator $\mathbb{Q}$.

### 4.3 Generative adversarial imitation learning

Recently, Ho and Ermon [12] proposed to directly learn a policy from data bypassing the intermediate IRL step. This work considers a specific case where the policy $\pi$ is a probability distribution over the state and actions, *i.e.*, $a_t \sim \pi(\cdot|s_t)$. They focus on the following optimization problem:

$$\min_\pi D_{JS}(\tilde{\rho}, \rho_\pi) - \lambda H(\pi) \tag{11}$$

where $\rho_\pi : \mathcal{S} \times \mathcal{A} \to \mathbb{R}$ is an occupancy measure induced by the policy $\pi$, which is defined as $\rho_\pi(s,a) = \pi(a|s)\sum_{t=0}^\infty P(s_t = s|\pi)$. This occupancy measure can be treated as the unnormalized distribution of state-action pairs.

The existence of the entropy term $H(\pi)$ is due to the assumption that the uncontrolled dynamics has a uniform distribution, since its analysis is based on the MaxEnt work. Another key aspect of this method is that it requires the occupancy measure $\rho_\pi$ has to be expressed directly using the policy $\pi$, which is applicable to MDPs. However, in point processes, stochastic differential equations or more complicated stochastic dynamics, there is no such simple or closed form expression between the probability measure and the policy. MDP is a simple case where the measure can be explicitly expressed in terms of the stochastic policy $\pi$. Hence this method is not applicable to SDEs or point processes for user behavior modeling.

## 5 Extensions

In this section, we discuss several extensions of our framework.

**Directly learning the policy**. Our framework can be extended and directly learn a policy $\pi$, bypassing the intermediate step of learning the cost function. Since we do not need to learn the cost function, we bypass the first step in our framework, which aims at finding $\mathbb{Q}^*_\theta$. Instead, we directly solve the following objective function:

$$\min_\pi D_W(\widetilde{\mathbb{Q}}||\mathbb{Q}_\pi) + \lambda D_W(\mathbb{Q}_\pi||\mathbb{P})$$

This objective captures the fact that we aim at learning a policy $\pi$, and keeping the measure $\mathbb{Q}_\pi$ that is induced by this policy, to be close to the baseline measure $\mathbb{P}$. Compared with the formulation that is based on MaxEnt IOC in (11), our formulation generalizes (11) to Wasserstein distance and incorporate the baseline measure $\mathbb{P}$.

**Learning both the policy and cost**. We can also jointly learn the state cost as well as the policy. We keep the two-step procedure in our Wasserstein framework, and only need to modify the second step by adding an extra regularization term $D_W(\mathbb{Q}^*_\theta||\mathbb{Q}_\pi)$. Mathematically, we modify (8) as follows:

$$\min_{\theta,\pi} D_W(\widetilde{\mathbb{Q}}||\mathbb{Q}^*_\theta) + \lambda D_W(\mathbb{Q}^*_\theta||\mathbb{Q}_\pi)$$

This objective function states that when computing $\theta$, we also try to find $\mathbb{Q}_\pi$ that is close to $\mathbb{Q}^*_\theta$.

# 6 Experiments

We evaluate our Wasserstein Deep IOC framework in two real world applications: the recommender system setting, where users will interact with different items at different times in online service platforms, and the social network setting, where users have different cost functions when they post messages in social platforms.

**Competitors**. We compare our Wasserstein inverse optimal control (W-IOC) with the following baselines. **KL-IOC** is our KL framework described in Section 2. **GAIL** is the generative adversarial imitation learning method [12]. **BC** is the behavior cloning method, and we use supervised learning to train the users' policy, using Adam [15] with minibatch of 128 samples.

Besides the IOC methods, to illustrate the difference between our work that treating users as RL agents and previous works that aim at fitting a generative model for users' behavior, we also compare with two state of the art methods in predicting users temporal behavior. **Coevolve** is a point process model that captures the evolution of users and items feature as the interaction event happens [26]. **Hawkes** is a low rank Hawkes process based generative model and it assumes user-item interactions are independent [6].

We use all algorithms to train policies of the same neural network for all applications: with two hidden layers of 100 unites each, and tanh nonlinearities in between.

## 6.1 Item recommendation in service platforms

Our datasets are obtained from three different domains from the TV streaming services (IPTV), the commercial review website (Yelp) and the online media services (Reddit). **IPTV** contains 7,100 users' watching history of 436 TV programs in 11 months, with 2,392,010 events, and 1,420 movie features, including 1,073 actors, 312 directors, 22 genres, 8 countries and 5 years. **Yelp** contains reviews for various businesses from October, 2004 to December, 2015. We filter users with more than 100 posts and it contains 100 users and 17,213 businesses with around 35,093 reviews. **Reddit** contains the discussions events in January 2014. Furthermore, we randomly selected 1,000 users and collect 1,403 groups that these users have discussion in, with a total of 10,000 discussion events. We use $p$ portion of the trajectories as the training data, and the rest as testing data. We vary the proportion $p \in \{0.5, 0.6, 0.7, 0.8\}$ and report the averaged results over five runs.

For each user, we learn his true cost function using our framework. Then we compute the optimal policy (choosing which item) using this cost function. We rank all items according to their score from the policy. Ideally, the each test item should have rank one. We repeat the evaluation on each testing item and report the TOP1 success rate of correcting predicting the items across all users. The success rate is computed as the proportion of times that the test item has rank one. Similarly, we report the TOP5 success rate, which is the proportion of times that the test item is within the top five items in the predicted item list.

Figure 1 shows that our W-IOC frameworks consistently performs the best across three different datasets, with the highest TOP1 and TOP5 success rate. Compared with GAIL and KL-IOC, our method has 10-20% higher success rate on average. This further shows the importance and benefit of using Wasserstein distance in our two-step framework, as compared with the JS distance in GAIL and KL distance in KL-IOC.

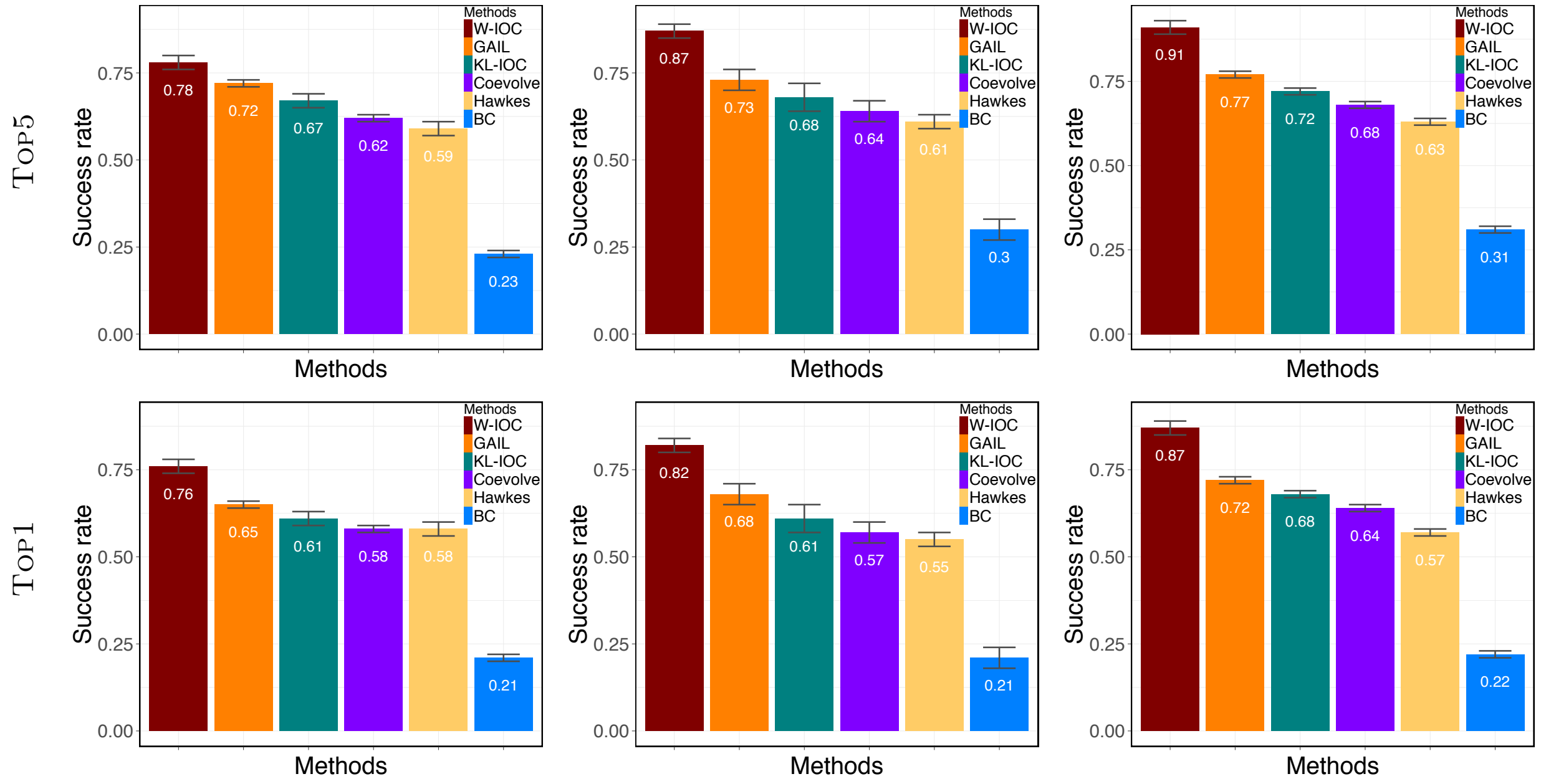

(a) IPTV (b) Yelp (c) Reddit

Figure 1: Prediction results on IPTV, Reddit and Yelp. Results are averaged over five runs with different portions of training trajectories and error bar represents the variance.

Compared with the two point process based models that have fixed parametric assumption of users' behavior patterns, our method also outperforms consistently. This highlight that our variational principle (viewing users as RL agents, and aiming at understanding the users' true cost function to predict future behavior) is both important and more expressive than the traditional phenomenological principle (fitting a generative model for users' temporal behavior).

## 6.2 Posting time prediction in social networks

We evaluate the performance of our framework on a real-world Twitter dataset, which contains 280,000 users with 550,000 tweets and retweets events. For each user, we track down all followers and record all the tweets they posted and reconstruct followers' timelines by collecting all the tweets by people they follow. We split the data in the same way as in in the recommendation system setting.

We treat each user as a RL algorithm and his policy is to decide whether to make a post in each time step. Given the observed trajectories, we learn the cost function, and compute the user's policy. In each test event, we rank all the user according to the score from the policy. The user that makes the post at that test time should have rank one. We define the TOP1 success rate as the proportion of times that the test user has rank one. Similarly, the TOP5 success rate is the proportion of times that the test user is within the top five users in the predicted user list.

Figure 2 further illustrates that our framework performs best. Note that the success rate in this task is lower than that in the recommendation setting, which suggests that in this task is harder than the recommender system task. This is because at each test event, we are predicting the top rank user among 280,000 users. Moreover, we also discovered that 90% of the Twitter user in this dataset are influenced by the behavior of his friends in deciding when to make a post, and the rest are more self-centric in making decisions to post.

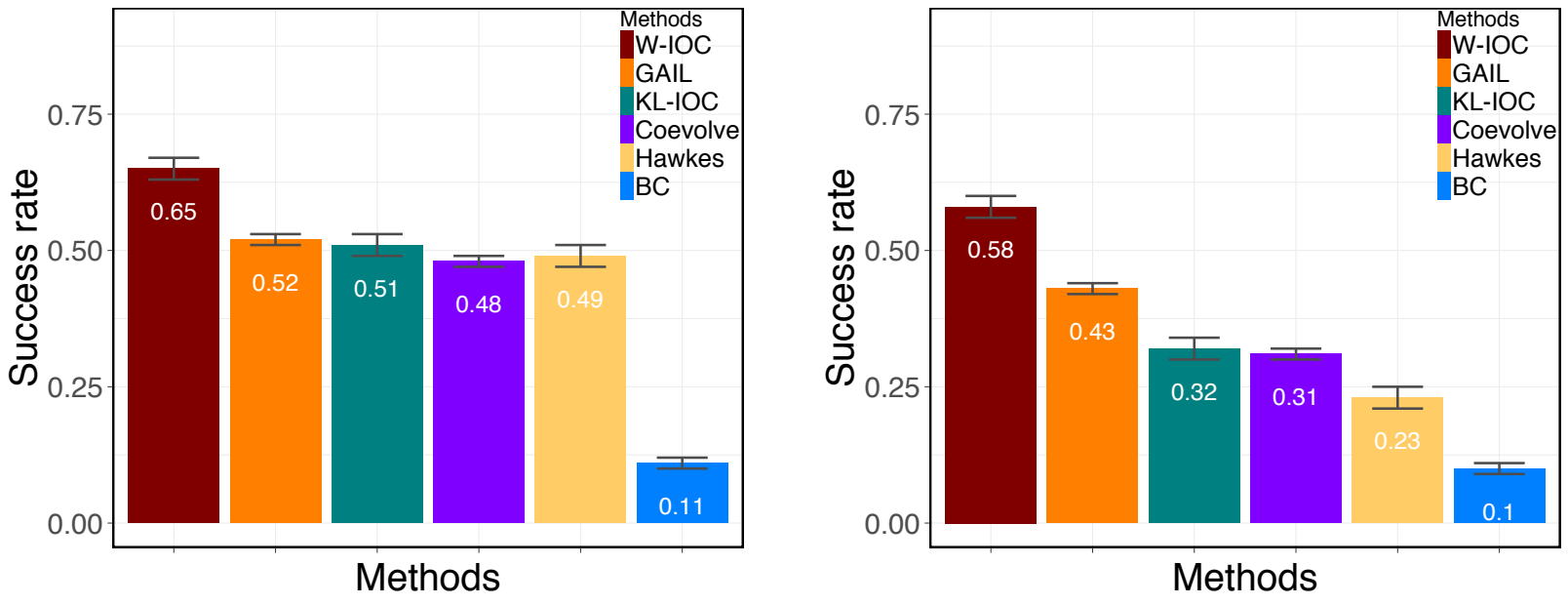

(a) Top5 success rate (b) Top1 success rate

Figure 2: Prediction results on Twitter dataset. Results are averaged over five runs with different portions of training trajectories and error bar represents the variance.

# 7 Conclusions

We have proposed a novel two-step Wasserstein deep IOC framework. In the first step, we compute the optimal measure using a mass transport equation. In the second step, we formulate the learning problem as a solving a generative adversarial network. We also presented a principled KL inverse optimal control framework that generalizes existing maximum entropy based works. In real world experiments, we show that our method achieves superior performance over existing IOC methods and generative models for user modeling.

In contrast to existing works in IOC, our framework does not require the analytical form for the underline distribution baseline $\mathbb{P}$, and only need the samples from $\mathbb{P}$ to compute the optimal measure $\mathbb{Q}_\theta^*$. Moreover, our framework does not assume a uniform dynamics for the baseline measure $\mathbb{P}$, and is applicable to general stochastic processes that includes MDPs as a special case. There are also many interesting directions for future work. For example, combined with the idea of learning an intermediate distribution in [9], our framework can also be generalized to the scenario where no information of the baseline measure or underline dynamics is available.

In the area of user behavior modeling, our work introduces a new variational principle: viewing users as RL algorithms, and aiming at understanding users' true cost function. This novel view is in sharp contrast to existing works [6, 17, 26, 31] that focus on fitting a generative model for the behavior data.

# A Proof of Theorem 1

The Wasserstein distance is defined as

$$D_W(\mathbb{Q}||\mathbb{P}) = \inf_{\Pi \in u(\mathbb{Q},\mathbb{P})} \mathbb{E}_{(x,y)\sim\Pi} d(x,y) \tag{12}$$

where $u(\mathbb{P},\mathbb{Q})$ denotes the set of all joint probability distributions $\Pi(x,y)$ whose marginals are respectively $\mathbb{P}$ and $\mathbb{Q}$. Note $\Pi(x,y)$ indicates how much mass must be transported from $x$ to $y$ in order to transform the distributions $\mathbb{P}$ and $\mathbb{Q}$.

Using the law of total probability, which asserts that any joint probability $\Pi$ can be constructed from marginal distribution $\mathbb{P}$ of $\tau'$ and the distribution $\mathbb{Q}_i$ of $\tau$ given $\tau' = \tau_i^0$, we can have the following expression for $\Pi$:

$$\Pi = \frac{1}{n}\sum_i \delta_{\tau_i^0} \otimes \mathbb{Q}_i$$

where each $\mathbb{Q}_i$ is a Dirac measure. Hence, we can express $\mathbb{Q}$ in the optimization problem as

$$\mathbb{Q} = \frac{1}{n}\sum_i \mathbb{Q}_i$$

With the structure of $\mathbb{Q}$, we can factorize the optimization problem in (5) as $n$ independent optimization problems:

$$\min_{\mathbb{Q}_i} \frac{1}{n}\sum_i \int c(\theta,\tau)\mathbb{Q}_i(\mathrm{d}\tau_i^0) + \gamma\frac{1}{n}\sum_i \int d(\tau,\tau_i^0)\mathbb{Q}_i(\mathrm{d}\tau)\ ,\ \text{for each } i \in [n]$$

Note that since $\mathbb{Q}_i$ is a Dirac measure, we can further simplify this optimization problem to find an optimal trajectory $\tau_i$ that solves the following optimization problem:

$$\min_{\tau_i} c(\theta,\tau) + \gamma d(\tau,\tau_i^0), \quad \text{for each } i \in [n] \tag{13}$$

The optimal solution $\tau_i^*$ to (13) uniquely defines the Dirac measure $\mathbb{Q}_i^*$.

# B Derivation of the Optimal Measure in (2)

The problem of finding the optimal measure is as follows:

$$\min_{\mathbb{Q}} \Big[\mathbb{E}_{\mathbb{Q}}[c(\theta,\tau)] + \gamma \mathbb{D}_{KL}(\mathbb{Q}||\mathbb{P})\Big],\ \text{s.t.} \int \mathrm{d}\mathbb{Q} = 1 \tag{14}$$

The minimum in (14) is attained at optimal measure $\mathbb{Q}^*$ given by:

$$\frac{\mathrm{d}\mathbb{Q}^*}{\mathrm{d}\mathbb{P}} = \frac{\exp(-\frac{1}{\gamma}c(\theta,\tau))}{\mathbb{E}_{\mathbb{P}}[\exp(-\frac{1}{\gamma}c(\theta,\tau))]} \tag{15}$$

Next, we show the derivations of (15), which contain two parts. First, we will show the following inequality:

$$\gamma \log\left(\mathbb{E}_{\mathbb{P}}\Big[\exp\big(-\frac{1}{\gamma}c(\theta,\tau)\big)\Big]\right) \leqslant \Big[\mathbb{E}_{\mathbb{Q}}[c(\theta,\tau)] + \gamma\mathbb{D}_{KL}(\mathbb{Q}||\mathbb{P})\Big] \tag{16}$$

The second part is to show the minimum of the above inequality is reached at (15).

To prove the first part, we first express $\mathbb{E}_{\mathbb{P}}$ in the left-hand-side of (16) as a function of the expectation $\mathbb{E}_{\mathbb{Q}}$. More specifically, we have:

$$\log\Big(\mathbb{E}_{\mathbb{P}}\Big[\exp\big(-\frac{1}{\gamma}c(\theta,\tau)\big)\Big]\Big) = \log\Big(\int \exp\big(-\frac{1}{\gamma}c(\theta,\tau)\big)\mathrm{d}\mathbb{P}\Big) \tag{17}$$

$$= \log\Big(\int \exp\big(-\frac{1}{\gamma}c(\theta,\tau)\big)\frac{\mathrm{d}\mathbb{P}}{\mathrm{d}\mathbb{Q}}\mathrm{d}\mathbb{Q}\Big) \tag{18}$$

$$\geqslant \int \log\Big(\exp\big(-\frac{1}{\gamma}c(\theta,\tau)\big)\frac{\mathrm{d}\mathbb{P}}{\mathrm{d}\mathbb{Q}}\Big)\mathrm{d}\mathbb{Q} \tag{19}$$

where (19) is due to the Jensen's inequality that puts the log operator inside the integral. The measure $\mathbb{P}$ is absolute continuous with respect to $\mathbb{Q}$, hence the derivative $\frac{\mathrm{d}\mathbb{P}}{\mathrm{d}\mathbb{Q}}$ exists.

Moreover, using the property that $\log(ab) = \log a + \log b$ and $\log(1/a) = -\log a$, the right-hand-side of the above inequality can be written as:

$$\begin{aligned}\int \log\Big(\exp\big(-\frac{1}{\gamma}c(\theta,\tau)\big)\frac{\mathrm{d}\mathbb{P}}{\mathrm{d}\mathbb{Q}}\Big)\mathrm{d}\mathbb{Q} &= \int\Big(-\frac{1}{\gamma}c(\theta,\tau) + \log\frac{\mathrm{d}\mathbb{P}}{\mathrm{d}\mathbb{Q}}\Big)\mathrm{d}\mathbb{Q}\\
&= \int -\frac{1}{\gamma}c(\theta,\tau)\mathrm{d}\mathbb{Q} + \int \log\frac{\mathrm{d}\mathbb{P}}{\mathrm{d}\mathbb{Q}}\mathrm{d}\mathbb{Q}\\
&= \int -\frac{1}{\gamma}c(\theta,\tau)\mathrm{d}\mathbb{Q} - \int \log\frac{\mathrm{d}\mathbb{Q}}{\mathrm{d}\mathbb{P}}\mathrm{d}\mathbb{Q}\\
&= -\frac{1}{\gamma}\mathbb{E}_{\mathbb{Q}}[c(\theta,\tau)] - \mathbb{D}_{\mathbb{KL}}(\mathbb{Q}||\mathbb{P}) \qquad (20)\end{aligned}$$

Hence, combining (19) and (20), we have:

$$\log\Big(\mathbb{E}_{\mathbb{P}}\Big[\exp\big(-\frac{1}{\gamma}c(\theta,\tau)\big)\Big]\Big) \geqslant -\frac{1}{\gamma}\mathbb{E}_{\mathbb{Q}}[c(\theta,\tau)] - \mathbb{D}_{\mathbb{KL}}(\mathbb{Q}||\mathbb{P}) \tag{21}$$

Finally, since $\gamma > 0$, multiply both sides of (21) by $-\gamma$ yields:

$$-\gamma\log\Big(\mathbb{E}_{\mathbb{P}}\Big[\exp\big(-\frac{1}{\gamma}c(\theta,\tau)\big)\Big]\Big) \leqslant \mathbb{E}_{\mathbb{Q}}[c(\theta,\tau)] + \gamma\mathbb{D}_{\mathbb{KL}}(\mathbb{Q}||\mathbb{P}) \tag{22}$$

This finishes the proof of (16), the first part of the theorem. Next, we will show the minimum is reached at $\mathbb{Q}^*$ given by (15).

To prove the second part, we will substitute (15) to the right-hand-side of (21) to show that the infimum is reached with this $\mathbb{Q}^*$. More specifically,

$$\begin{aligned}&\mathbb{E}_{\mathbb{Q}^*}[c(\theta,\tau)] + \gamma\mathbb{D}_{\mathbb{KL}}(\mathbb{Q}^*||\mathbb{P}) \qquad (23)\\
&= \mathbb{E}_{\mathbb{Q}^*}[S(\boldsymbol{x})] + \gamma\int\log\frac{\mathrm{d}\mathbb{Q}^*}{\mathrm{d}\mathbb{P}}\mathrm{d}\mathbb{Q}^*\\
&= \mathbb{E}_{\mathbb{Q}^*}[c(\theta,\tau)] + \gamma\int\log\frac{\exp(-\frac{1}{\gamma}c(\theta,\tau))}{\mathbb{E}_{\mathbb{P}}[\exp(-\frac{1}{\gamma}c(\theta,\tau))]}\mathrm{d}\mathbb{Q}^*\\
&= \mathbb{E}_{\mathbb{Q}^*}[c(\theta,\tau)] + \gamma\int -\frac{1}{\gamma}c(\theta,\tau)\mathrm{d}\mathbb{Q}^* - \gamma\int\log\Big(\mathbb{E}_{\mathbb{P}}\Big[\exp\big(-\frac{1}{\gamma}c(\theta,\tau)\big)\Big]\Big)\mathrm{d}\mathbb{Q}^* \qquad (24)\\
&= \mathbb{E}_{\mathbb{Q}^*}[c(\theta,\tau)] - \int c(\theta,\tau)\mathrm{d}\mathbb{Q}^* - \gamma\log\Big(\mathbb{E}_{\mathbb{P}}\Big[\exp\big(-\frac{1}{\gamma}c(\theta,\tau)\big)\Big]\Big)\int\mathrm{d}\mathbb{Q}^*\\
&= \mathbb{E}_{\mathbb{Q}^*}[c(\theta,\tau)] - \mathbb{E}_{\mathbb{Q}^*}[c(\theta,\tau)] - \gamma\log\Big(\mathbb{E}_{\mathbb{P}}\Big[\exp\big(-\frac{1}{\gamma}c(\theta,\tau)\big)\Big]\Big) \qquad (25)\\
&= -\gamma\log\Big(\mathbb{E}_{\mathbb{P}}\Big[\exp\big(-\frac{1}{\gamma}c(\theta,\tau)\big)\Big]\Big)\end{aligned}$$

where (24) is due to the property $\log(a/b) = \log a - \log b$ and (25) is because $\mathbb{Q}^*$ is a probability measure hence $\int d\mathbb{Q}^* = 1$. Hence the infimum is reached and this finishes the proof of the second part.

# C Proof of Theorem 2

To prove Theorem 2, we first present a general lemma as follows.

**Lemma 3.** *Let $g : \mathbb{R} \times \mathbb{R} \to \mathbb{R}$ be a continuous function with first and second derivatives. Let $h(x) = \operatorname{argmin}_y g(x, y)$. Then we have:*

$$\frac{dh(x)}{dx} := h'(x) = \Big[\frac{\partial^2 g(x, h(x))}{\partial y^2}\Big]^{-1} \frac{\partial^2 g(x, h(x))}{\partial x \partial y} \tag{26}$$

*Proof.* Since $h(x) = \operatorname{argmin}_y g(x, y)$, we have

$$\frac{\partial g(x, y)}{\partial y}\Big|_{y=h(x)} = 0$$

Differentiating both right hand side and left hand side, we have:

$$\frac{d}{dx} \frac{\partial g(x, y)}{\partial y} = 0$$

Using chain rule, we also have that

$$\frac{d}{dx} \frac{\partial g(x, y)}{\partial y} = \frac{\partial^2 g(x, h(x))}{\partial xy} + \frac{\partial^2 g(x, h(x))}{\partial xy} g'(x)$$

Combing these two equations, we have (26). □

To prove Theorem 2, we just need to set $\tau(\cdot)$ as $h(\cdot)$.